\documentclass[journal,onecolumn]{IEEEtran}

\usepackage[english]{babel}
\usepackage[utf8]{inputenc}
\usepackage{amsmath}
\usepackage{graphicx}
\usepackage{xparse}
\usepackage{setspace}
\usepackage{array}
\usepackage[dvipsnames]{xcolor}
\usepackage{bbm}
\usepackage{enumitem}

\usepackage{commath}
\usepackage{bm}
\usepackage{algorithmic}
\usepackage[ruled]{algorithm2e}
\usepackage{array}
\usepackage{url}
\usepackage{hyperref}

\usepackage{etoolbox}
\makeatletter
\patchcmd{\@makecaption}
  {\scshape}
  {}
  {}
  {}
\makeatother

\title{\LARGE Federated Neuromorphic Learning of Spiking Neural Networks \\ for Low-Power Edge Intelligence}
\author{Nicolas Skatchkovsky, Hyeryung Jang, and Osvaldo Simeone \thanks{The authors are with the King's Centre for Learning and Information Processing (kclip), Department of Engineering, King's College London, United Kingdom. The authors have received funding from the European Research Council (ERC) under the European Union’s Horizon 2020 Research and Innovation Programme (Grant Agreement No. 725731). Code can be found at \url{https://github.com/kclip}.}}

\begin{document}
%
\maketitle
\begin{abstract}
Spiking Neural Networks (SNNs) offer a promising alternative to conventional Artificial Neural Networks (ANNs) for the implementation of on-device low-power online learning and inference. On-device training is, however, constrained by the limited amount of data available at each device. In this paper, we propose to mitigate this problem via cooperative training through Federated Learning (FL). To this end, we introduce an online FL-based learning rule for networked on-device SNNs, which we refer to as FL-SNN. FL-SNN leverages local feedback signals within each SNN, in lieu of backpropagation, and global feedback through communication via a base station. The scheme demonstrates significant advantages over separate training and features a flexible trade-off between communication load and accuracy via the selective exchange of synaptic weights. 
\end{abstract}
\section{Introduction}
\label{sec:intro}

 Training state-of-the-art Artificial Neural Network (ANN) models requires distributed computing on large mixed CPU-GPU clusters, typically over many days or weeks, at the expense of massive memory, time, and energy resources \cite{DBLP:journals/corr/abs-1906-02243}, and potentially of privacy violations. Alternative solutions for low-power machine learning on resource-constrained devices have been recently the focus of intense research, and this paper proposes and explores the convergence of two such lines of inquiry, namely Spiking Neural Networks (SNNs) for low-power on-device learning \cite{loihi, Fouda2019SpikingNN, Nandakumar2019SupervisedLI, intro_snn}, and Federated Learning (FL) \cite{FL_paper,federated_learning} for collaborative inter-device training.

SNNs are biologically inspired neural networks in which neurons are dynamic elements processing and communicating via sparse spiking signals over time, rather than via real numbers,  enabling the native processing of time-encoded data, e.g., from DVS cameras \cite{dvs_camera}. They can be implemented on dedicated hardware \cite{loihi}, offering energy consumptions as low as a few picojoules per spike \cite{intro_snn}.

With FL, distributed devices can carry out collaborative learning without exchanging local data. This makes it possible to train more effective machine learning models by benefiting from data at multiple devices with limited privacy concerns. FL requires devices to periodically exchange information about their local model parameters through a parameter server \cite{federated_learning}.    

\begin{figure}[t]
    \centering
    \includegraphics[scale=0.3]{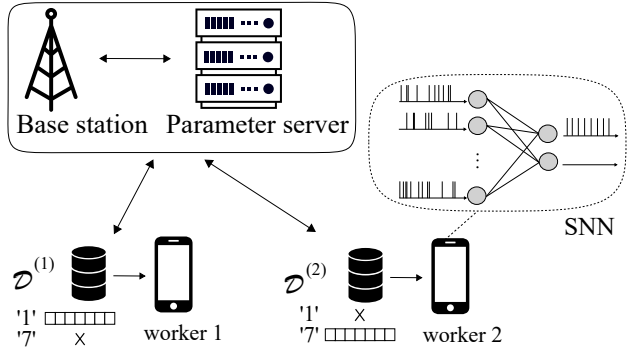}
    \caption{Federated Learning (FL) model under study: Mobile devices collaboratively train on-device SNNs based on different, heterogeneous, and generally unbalanced local data sets, by communicating through a base station (BS).}
    \label{system_model}
\end{figure}

This paper explores for the first time the implementation of collaborative FL for on-device SNNs. As shown in Fig.~\ref{system_model}, we consider a number of mobile devices, each training a local SNN on the basis of local data, which can communicate through a base station (BS) to implement FL. The proposed FL solution enables learning and inference on time-encoded data, and enables a flexible trade-off between communication load and accuracy. It generalizes the single-SNN learning rule introduced in \cite{intro_snn} for probabilistic SNNs, which in turn recovers as special cases many known training algorithms for SNNs such as Spike-Timing-Dependent Plasticity (STDP) and INST/FILT \cite{gardner16INST}.
\vspace{5pt}

\section{System Model} \label{sec:model}

As illustrated in Fig.~\ref{system_model}, we consider a distributed edge computing architecture in which $N$ mobile devices communicate through a BS in order to perform the collaborative training of local SNN models via FL. Each device $i = 1, \dots, N$ holds a different local data set $\mathcal{D}^{(i)}$ that contains a collection of $|\mathcal{D}^{(i)}|$ data points $\{z_{j}\}_{j=1}^{|\mathcal{D}^{(i)}|}$. Each sample $z_{j}$ is given by a pair $z_{j} = (x_{j}, y_{j})$ of covariate vector $x_{j}$ and desired output $y_{j}$. The goal of FL is to train a common SNN-based model without direct exchange of the data from the local data sets. In the rest of this section, we first review conventional FL and then separate online training of individual SNNs \cite{intro_snn}.

\vspace{-0.2cm}
\subsection{Conventional Federated Learning} \label{sec:FL}
Denoting as $f(\mathbf{\theta}, z)$ the loss function for any example $z$ when the model parameter vector is $\theta$, the local empirical loss at a device $i$ is defined as 
\begin{equation}
F^{(i)}(\mathbf{\theta}) = \frac{1}{|\mathcal{D}^{(i)}|}\sum_{z \in \mathcal{D}^{(i)}} f(\mathbf{\theta}, z).
\end{equation}
The global learning objective of conventional FL is to solve the problem
\begin{equation}
    \label{general_objective}
     \min_{\theta} F(\mathbf{\theta}) := \frac{1}{\sum_{i=1}^{N}  |\mathcal{D}^{(i)}|}\sum_{i=1}^{N}|\mathcal{D}^{(i)}| F^{(i)}(\mathbf{\theta}),
\end{equation}
where $F(\theta)$ is the global empirical loss over the collective data set $\cup_{i=1}^{N} \mathcal{D}^{(i)}$.  FL proceeds in an iterative fashion across $T$ iterations, with one communication update through the BS every $\tau$ iterations (see, e.g., \cite{FL_paper}). To elaborate, at each iteration $t = 1, \dots, T$, each device $i$ computes a local Stochastic Gradient Descent (SGD) update
\begin{equation}
    \label{general_learning_rule}
    \mathbf{\theta}^{(i)}(t) = \tilde{\theta}^{(i)}(t-1) - \alpha \nabla f \big( \tilde{\theta}^{(i)}(t-1), z^{(i)}(t) \big),
\end{equation}
where $\alpha$ is the learning rate; $\tilde{\theta}^{(i)}(t-1)$ represents the local value of the model parameter at the previous iteration $t - 1$ for device $i$; and $z^{(i)}(t)$ is a randomly selected example (with replacement) from local data set $\mathcal{D}^{(i)}$. If $t$ is not a multiple of $\tau$, we set $\tilde{\mathbf{\theta}}^{(i)}(t) = \theta^{(i)}(t)$ for all devices $i = 1, \dots, N$. Otherwise, when $t = \tau, 2\tau, \dots$, the devices communicate the updated parameter \eqref{general_learning_rule} to the BS, which computes the centralized averaged parameter
\begin{equation}
\label{global_update}
    \theta(t) = \frac{1}{\sum_{i=1}^{N} |\mathcal{D}^{(i)}|} \sum_{i=1}^{N} |\mathcal{D}^{(i)}|\theta^{(i)}(t).
\end{equation}
The updated parameter \eqref{global_update} is sent back to all devices via multicast downlink transmission. 
This is used in the next iteration $t+ 1$ as initial value $\tilde{\theta}^{(i)}(t) = \theta(t)$ for all $i = 1, \dots, N$.

\subsection{Probabilistic SNN Model and Training
}
\label{ondevice_snns}

\begin{figure}[t]
    \centering
    \includegraphics[scale=0.7]{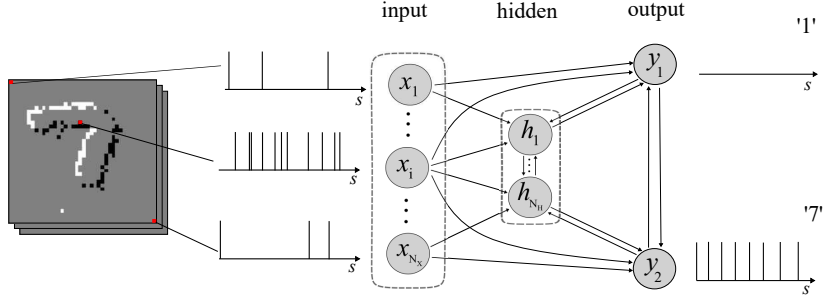}
    \caption{Example of an internal architecture for an on-device SNN with $N_X$ input, $N_Y = 2$ output, and $N_H$ hidden spiking neurons.}
    \label{snn_architecture}
\end{figure}

An SNN is a network of $N_{V}$ spiking neurons connected via an arbitrary directed graph, possibly with cycles (see Fig.~\ref{snn_architecture}). Each neuron $n \in \mathcal{V}$ receives the signals emitted by the subset $\mathcal{P}_{n}$ of neurons connected to it through directed links, known as \textit{synapses}. At any local algorithmic time $s = 1, \dots, S$, each neuron $n$ outputs a binary signal $o_{n}(s) \in \{0, 1\}$, with ``$1$'' representing a spike.
As seen in Fig.~\ref{snn_architecture}, neurons in the SNN can be partitioned into three disjoint subsets as $\mathcal{V} = \mathcal{X} \cup \mathcal{Y} \cup \mathcal{H}$, with $\mathcal{X}$ being the subset of \textit{input} neurons, $\mathcal{Y}$ the subset of \textit{output} neurons, and $\mathcal{H}$ the subset of \textit{hidden}, or \textit{latent}, neurons.

Following the probabilistic Generalized Linear Model (GLM), the spiking behavior of any neuron $n \in \mathcal{V}$ at local algorithmic time $s$ is determined by its \textit{membrane potential} $u_{n}(s)$: The instantaneous spiking probability of neuron $n$ at time $s$, conditioned on the value of the membrane potential, is 
\begin{align}
    \label{spiking_proba}
    & p\big( o_{n}(s)=1 | u_{n}(s) \big) = 
    \sigma \big( u_{n}(s) \big),
\end{align}
with $\sigma(\cdot)$ being the sigmoid function \cite{intro_snn}.

The membrane potential depends on the previous spikes of pre-synaptic neurons $\mathcal{P}_n$ and on the past spiking behavior of the neuron itself, according to the formula
\begin{equation} \label{membrane_potential}
    u_{n}(s) = \sum_{k \in \mathcal{P}_{n}} \overrightarrow{o}_{n, k}(s-1) + w_{n}\overleftarrow{o}_{n}(s-1) + \gamma_{n},
\end{equation}
where  $\gamma_{n}$ is a bias parameter; the contribution of each pre-synaptic neuron $k \in \mathcal{P}_{n}$ is through the filtered synaptic trace $\overrightarrow{o}_{n,k }(s) = a_{n,k}(s)*o_{k}(s)$, and that of the neuron itself through its feedback trace $\overleftarrow{o}_{n}(s) =  b(s) * o_{n}(s)$, where $*$ denotes the convolution operator; and $\{ a_{n,k}(s) \}_{k \in \mathcal{P}_{n}}$ and $b_{n}(s)$ are the synaptic and feedback impulse responses. Each impulse response $a_{n, k}(s)$ is given by the linear combination $a_{n,k}(s) = \sum_{l=1}^{K_{a}}w_{n,k}^{l}a_{l}(s)$
with $K_a$ fixed basis functions $\{a_{l}(s)\}_{l=1}^{K_a}$ and $K_a$ \textit{learnable} synaptic feedforward weights $\{w_{n, k}^{l}\}_{l=1}^{K_a}$, whereas the feedback filter $b(s)$ is fixed with a learnable feedback weight $w_n$. 

During the (individual) training of an SNN, for each selected example $z = (x, y)$, the covariate vector $x$ is first encoded into input binary time-sequences $\{ \mathbf{x}(s) \}_{s=1}^{S'} = \{ x_{n}(s) : n \in \mathcal{X} \}_{s=1}^{S'}$ and the target $y$ into desired output binary time-sequences $\{ \mathbf{y}(s) \}_{s=1}^{S'} = \{y_{n}(s) : n \in \mathcal{Y} \}_{s=1}^{S'}$ of duration $S'$ samples \cite{intro_snn}. At each local algorithmic time $s$, the spike signal of each input neuron $n \in \mathcal{X}$ is assigned value $x_{n}(s)$; while each output neurons $n \in \mathcal{Y}$ is assigned value $y_{n}(s)$. 
The negative log-probability of obtaining the desired output sequence $\mathbf{y}$ given input $\mathbf{x}$ is computed by summing over all possible values of the latent spike signals $\mathbf{h}$ as
\begin{align}
 \label{log-likelihood}
 \mathcal{L}_{\mathbf{y} | \mathbf{x}}(\theta) &= - \log \sum_{\mathbf{h}} \prod_{s=1}^S \prod_{n \in \mathcal{Y} \cup \mathcal{H}} p\big( o_{n}(s) | u_{n}(s) \big),
\end{align}
where the vector $\mathbf{\theta} = \{ \gamma_{n}, \{\{ w_{k,n}^{l}\}_{l=1}^{K_a}\}\}_{k \in \mathcal{P}_{n}}, w_{n} \}_{n \in \mathcal{V}}$ is the model parameters of an SNN. When interpreted as a function of $\theta$, the quantity \eqref{log-likelihood} is known as negative log-likelihood or \textit{log-loss}. When summed over the available data set, the log-loss can be optimized as a learning criterion via SGD \cite{intro_snn}.
\vspace{5pt}

\section{FL-SNN: FL with Distributed SNNs}
\label{sec:fl_snn}
In this section, we propose a novel online FL algorithm, termed FL-SNN, that jointly trains on-device SNNs. FL-SNN aims at minimizing the global loss \eqref{general_objective}, where the local loss function is given by the log-loss \eqref{log-likelihood} as
\begin{equation} 
\label{SNN_local_loss}
 F^{(i)} \big( \theta \big) = \frac{1}{|\mathcal{D}^{(i)}|} \sum_{z = (x, y) \in \mathcal{D}^{(i)}} \mathcal{L}_{\mathbf{y} | \mathbf{x}}(\theta).
\end{equation}
In \eqref{SNN_local_loss}, $\theta$ is the set of learnable parameters for a common SNN-based model defined by \eqref{spiking_proba}-\eqref{membrane_potential}; 
and  $(\mathbf{x}, \mathbf{y})$ are the input and output binary time-sequences of length $S'$ encoding example $z=(x,y) \in \mathcal{D}^{(i)}$.

To start, each device $i$ selects a sequence of $D$ examples from the local training set $\mathcal{D}^{(i)}$, uniformly at random (with replacement). Concatenating the examples yields time sequences $(\mathbf{x}^{(i)}, \mathbf{y}^{(i)})$ of $S$ binary samples, with $S = DS'$. We note that, between any two examples, one could include an interval of time, also known in neuroscience as inter-stimulus interval.  FL-SNN tackles problem \eqref{general_objective} via the minimization of the log-loss $\sum_{i=1}^{N} \mathcal{L}_{\mathbf{y}^{(i)} | \mathbf{x}^{(i)}}(\theta)$ through online gradient descent \cite{intro_snn} and periodic averaging \eqref{global_update}.

\vspace{10pt}
\begin{table}[h!]
\begin{center}
\begin{tabular}{ | m{1.2cm} || m{1.2cm}| m{1.2cm} | m{1.2cm} | m{1.2cm}| m{1.2cm} | m{1.2cm} | } 
 \hline
 $s$ & 1 & 2 & 3 & 4 & 5 & 6 \\
 \hline
 $t$  & \ & 1 & \ & 2 & \ & 3 \\
\hline
 $\tilde{\theta}^{(i)}$ & \textcolor{red}{$\theta(0)$} & \textcolor{blue}{$\theta^{(i)}(1)$} & $\theta^{(i)}(1)$ & \textcolor{red}{$\theta(2)$} & $\theta(2)$ & \textcolor{blue}{$\theta^{(i)}(3)$} \\
 \hline 
\end{tabular}
  \vspace{4pt}
  \caption{Illustration of the time scales involved in the cooperative training of SNNs via FL for $\tau = 2$ and $\Delta s = 2$. The third line indicates the values of the parameter at the end of local algorithmic steps at any device $i$ during training. Blue indicates local updates, while red indicates global updates.}
\label{time_scales}
\end{center}
\end{table}

\begin{algorithm}[t]
\label{algo:FL_SNN}
\DontPrintSemicolon
\SetAlgoLined
\KwIn{Parameters: $\tau, T, \Delta s$; Hyperparameters: $\alpha, \kappa$; Data: $(\mathbf{x}^{(i)}, \mathbf{y}^{(i)})_{i=1}^N$}
\KwOut{Final model parameters $\theta_{F}$}
\vspace{0.1cm}
\hrule
\vspace{0.1cm}
 {\bf initialize} $\theta^{(i)}(0) = \bar{\theta}(0)$ for all devices $i = 1, \dots, N$ \\ 
 \For{{\em all global algorithmic time steps $t = 1, \dots, T$}}{
    \For{{\em all devices $i = 1,\dots,N$}}{
    \For{{\em each hidden neuron
    $n \in \mathcal{H}$}}{
        \begin{eqnarray}
        h_{n}^{(i)}(s) \sim \text{Bern}(\sigma(u_{n}^{(i)}(s)))
        \end{eqnarray}
         sequentially for all $s \in [t]$ 
    } 
   compute the learning signal $\ell^{(i)}(t)$ as \eqref{learning_signal}

   \For{{\em each neuron $n \in \mathcal{H} \cup \mathcal{Y}$}}{
 local parameter $\theta^{(i)}(t)$ update \eqref{snn_update} with eligibility trace \eqref{egilibility_trace}
   }
\smallskip
  \eIf{$t \mod{\tau} = 0$}{
   transmit $\theta^{(i)}(t)$ to the BS  \\
   \smallskip
   fetch $\theta(t)$ in \eqref{global_update} from the BS  \\
   \smallskip
   set $\tilde{\theta}^{(i)}(t) = \theta(t)$
   }
   {
   set $ \tilde{\theta}^{(i)}(t) = \theta^{(i)}(t)$
  }
  }
 }
\caption{FL-SNN}
\end{algorithm}

The resulting FL-SNN algorithm is summarized in Algorithm \ref{algo:FL_SNN}. As illustrated in Table~\ref{time_scales}, each global algorithmic iteration $t$ corresponds to $\Delta s$ local SNN time steps, which we define as the interval $[t] = \{s: (t-1)\Delta s + 1 \leq s \leq t\Delta s \}$. Note that the total number $S$ of SNN local algorithmic time steps and the number $T$ of global algorithmic time steps during the training procedure are hence related as $S = DS' = T\Delta s$. Therefore, unlike the conventional FL, the number $D$ of examples is generally different from the number $T$ of local updates. 

At each global algorithmic iteration $t = 1, \ldots, T$, upon generation of spiking signals \eqref{spiking_proba} by the hidden neurons, the local update rule of the SNN at each neuron $n$ in the SNN of device $i$ is given by the online gradient steps for loss $\mathcal{L}_{\mathbf{y}^{(i)} | \mathbf{x}^{(i)}}(\theta)$ \cite{intro_snn}
\begin{equation}
\label{snn_update}
    \mathbf{\theta}^{(i)}_{n}(t) = \tilde{\theta}^{(i)}_{n}(t-1) - \alpha \cdot \begin{cases}
{e}^{(i)}_{n}(t) &\text{for  $n \in \mathcal{Y}$}\\
\ell^{(i)}(t){e}^{(i)}_{n}(t) &\text{for $n \in \mathcal{H}$}
\end{cases},
\end{equation}

with
\begin{equation}
    \ell^{(i)}(t)  = \kappa \ell^{(i)}(t-1)  + (1 - \kappa)\sum_{\substack{s \in [t], \\n \in \mathcal{X} \cup \mathcal{Y}}} \log p \big(o_{n}^{(i)}(s)|u_{n}^{(i)}(s) \big)
\label{learning_signal}
\end{equation}

and
\begin{flalign}
\label{egilibility_trace}
e_{n}^{(i)}(t)  = \kappa e_{n}^{(i)}(t-1)  +  (1 - \kappa) \sum_{s \in [t]} \nabla \log p\big( o_{n}^{(i)}(s) | u^{(i)}_{n}(s) \big),
\end{flalign}
which respectively correspond to the learning signal and eligibility trace, i.e., the running average of the gradients of the log-loss. The global update at the BS is then given by \eqref{global_update}. 

As summarized in Algorithm \ref{algo:FL_SNN}, FL-SNN is based on local and global feedback signals, rather than backpropagation. As in \cite{intro_snn}, the local learning signal $\ell^{(i)}(t)$, computed every $\Delta s$, indicates to the hidden neurons within the SNN of each device $i$ how effective their current signaling is in maximizing the probability of the desired input-output behavior defined by the selected data $(\mathbf{x}^{(i)}, \mathbf{y}^{(i)})$. In contrast, the global feedback signal $\theta(t)$ is given by the global averaged parameter \eqref{global_update}, which aims at enabling cooperative training via FL.

\vspace{5pt}

\section{Experiments}
\label{sec:experiments}
\begin{figure}[t]
    \centering
    \includegraphics[scale=.5]{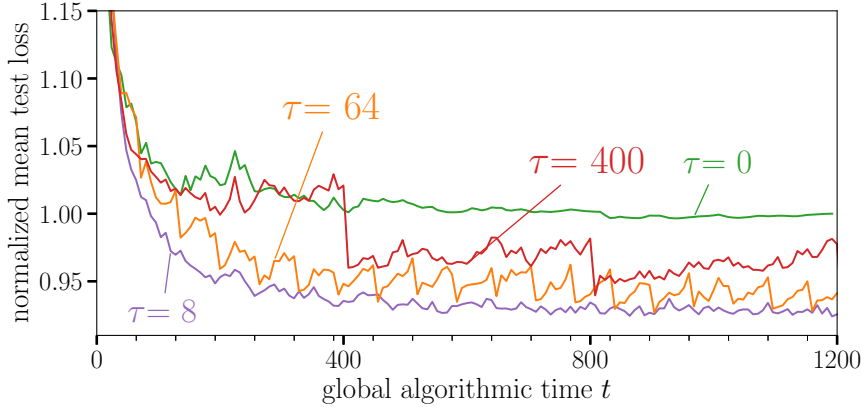} 
    \caption{Evolution of the mean test loss during training for different values of the communication period $\tau$ ($N_{H}=0$).}
    \label{fig_loss}
\end{figure}
\vspace{-5pt}

We consider a handwritten digit classification task based on the MNIST-DVS dataset \cite{mnist-dvs}.  The latter was obtained by displaying slowly moving handwritten digits from the MNIST dataset on an LCD monitor and recording the output of a $128 \times 128$ pixel DVS (Dynamic Vision Sensor) camera \cite{mnist, dvs_camera}. The camera uses send-on-delta encoding, whereby for each pixel, positive ($+1$) or negative ($-1$) events are recorded if the pixel's luminosity respectively increases or decreases by more than a given amount, and no event is recorded otherwise. Following  \cite{6933869,DBLP:journals/corr/HendersonGW15}, images are cropped to $26 \times 26$ pixels, which capture the active part of the image, and to $2$ seconds. Uniform downsampling over time is then carried out to obtain $S'= 80$ samples. The training dataset  
is composed of $900$ examples per class and the test dataset  
is composed of $100$ samples per class. 

We consider $N = 2$ devices which have access to disjoint subsets of the training dataset. In order to validate the advantages of FL, we assume that the first device has only samples from class `$1$' and the second only from class `$7$'. As seen in Fig.~\ref{snn_architecture}, each device is equipped with an SNN with directed synaptic links existing from all neurons in the input layer, consisting of $26 \times 26 = 676$ neurons, to all other neurons, consisting of $N_{H}$ hidden and $2$ output neurons. Hidden and output neurons are fully connected. We choose other network parameters as: $K_{a} = 8$ synaptic basis functions, selected as raised cosine functions with a synaptic duration of $10$ time-steps \cite{pillow}; and training parameters $\alpha = 0.05$, $\kappa = 0.2$, and $\Delta s = 5$. We train over $D = 400$ randomly selected examples from the local data sets, which results in $S = D S' = 32, 000$ local time-steps.

As a baseline, we consider the test loss at convergence for the separate training of the two SNNs. In Fig.~\ref{fig_loss}, we plot the local test loss normalized by the mentioned baseline as a function of the global algorithmic time $t$ for $N_{H}=0$. A larger communication period $\tau$ is seen to impair the learning capabilities of the SNNs, yielding a larger final value of the loss. In fact, e.g, for $\tau = 400$, after a number of local iterations without communication, the individual devices are not able to make use of their data to improve performance.

\begin{figure}[h]
    \centering
    \includegraphics[scale=.5]{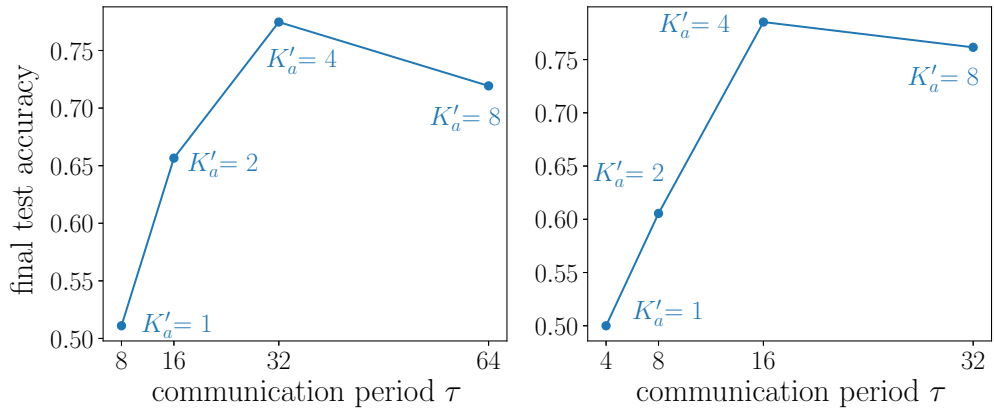} 
    \caption{Final test accuracies for different values of $\tau$ with a fixed communication rates $r = 1/8$ (left) and $r=1/4$ (right) of synaptic weights per global iteration ($N_{H} = 16$).}
    \label{fig:fixed_rate}
\end{figure}

As noted in recent literature \cite{federated_learning, sparse_ternary, signsgd, Aji_2017}, one of the major flaws of FL is the communication load incurred by the need to regularly transmit large model parameters. To partially explore this aspect, we now consider exchanging only a subset $K'_{a} \leq K_{a}$ of synaptic weights during global iterations. Define the communication rate as the number of synaptic weights exchanged per global iteration, i.e., $r = K'_{a}/\tau$. We assume that, for a given rate $r$ and period $\tau$, each device communicates only the $K'_{a} = r\tau$ weights with the largest gradients \eqref{snn_update}. The BS averages the weights sent by both devices; set the weights transmitted by one device only to the given transmitted value; and set unsent weights to zero. We note that the overhead of $1$ byte to communicate the position of the weights sent is small with respect to the overall communication load. In Fig. \ref{fig:fixed_rate}, we plot the final test accuracy as a function of $\tau$ for fixed rates $r=1/8$ and $r=1/4$ with $N_H = 16$. We observe that it is generally preferable to transmit a subset of weights more frequently in order to enhance cooperative training.

\vspace{5pt}

\section{Conclusion}
\label{sec:conclusion}
This paper introduced a novel protocol for communication and cooperative training of on-device Spiking Neural Networks (SNNs). We demonstrated significant advantages over separate training, and highlighted new opportunities to trade communication load and accuracy by exchanging subsets of synaptic weights.

\newpage

\bibliographystyle{IEEEbib}
\bibliography{biblio}

\end{document}